\documentclass[letterpaper, 10pt, conference]{ieeeconf}      

\IEEEoverridecommandlockouts                              

\usepackage{graphics} 
\usepackage{color}
\usepackage{subcaption}
\usepackage[british]{babel}
\usepackage{multirow}
\usepackage[hyphens]{url}
\usepackage{hyperref}
\usepackage{cite}
\usepackage{epsfig} 
\usepackage{amsmath,amssymb}
\usepackage{multirow}
\usepackage{algorithm,algorithmic}
\usepackage[font=small]{caption} 
\usepackage{subcaption}
\usepackage{multirow}
\usepackage{url}
\usepackage{comment}
\usepackage[table]{xcolor}

\graphicspath{ {./images/} }
\DeclareGraphicsExtensions{.png,.pdf,.jpeg,.jpg}

\usepackage{todonotes}
\usepackage{soul}
\definecolor{smoothgreen}{rgb}{0.7,1,0.7}
\sethlcolor{smoothgreen}

\definecolor{darkpink}{rgb}{0.91, 0.33, 0.5}

\newcommand {\vikram}[1]{{\color{black}#1}}

\title{\LARGE \bf Learning to Assess Danger from Movies for Cooperative Escape Planning in Hazardous Environments}

\author{Vikram Shree, Sarah Allen, Beatriz Asfora, Jacopo Banfi, and Mark Campbell
\thanks{V. Shree, S. Allen, B. Asfora, and M. Campbell are with the Sibley School of Mechanical and Aerospace Engineering, Cornell University, Ithaca, NY USA. Email: {\tt\small \{vs476, sea97, ba386, mc288\}@cornell.edu}. J. Banfi is with the Computer Science and Artificial Intelligence Laboratory, Massachusetts Institute of Technology, Cambridge, MA USA. Email: {\tt\small jbanfi@mit.edu}}
\thanks{Research supported by the NRI program of the National Science Foundation, award $\#$1830497.}
\thanks{This manuscript has been accepted for publication at 2022 IEEE/RSJ International Conference on Intelligent Robots and Systems (IROS).}}

\begin{document}

\maketitle
\begin{abstract}
    \label{sec:abstract}
    \vikram{There has been a plethora of work towards improving robot perception and navigation, yet their application in hazardous environments, like during a fire or an earthquake, is still at a nascent stage. We hypothesize two key challenges here: first, it is difficult to replicate such scenarios in the real world, which is necessary for training and testing purposes. Second, current systems are not fully able to take advantage of the rich multi-modal data available in such hazardous environments. To address the first challenge, we propose to harness the enormous amount of visual content available in the form of movies and TV shows, and develop a dataset that can represent hazardous environments encountered in the real world. The data is annotated with high-level danger ratings for realistic disaster images, and corresponding keywords are provided that summarize the content of the scene. In response to the second challenge, we propose a multi-modal danger estimation pipeline for collaborative human-robot escape scenarios. Our Bayesian framework improves danger estimation by fusing information from robot's camera sensor and language inputs from the human. Furthermore, we augment the estimation module with a risk-aware planner that helps in identifying safer paths out of the dangerous environment. Through extensive simulations, we exhibit the advantages of our multi-modal perception framework that gets translated into tangible benefits such as higher success rate in a collaborative human-robot mission.}
\end{abstract}


\section{Introduction}
\label{sec:introduction}



\vikram{In the past decade, there has been a surge in the application of robotics in different avenues of our day-to-day life. Think, for example, of self-driving cars, cleaning robots, and robots as personal assistants. A few of these robots have achieved remarkable success while operating in organized environments with limited uncertainty like factories and homes. 
Yet, the deployment of robots during the World Trade Center disaster \cite{casper2003human} and a more recent application during the Surfside condominium collapse \cite{murphy2021how}, revealed significant untapped potential of current robotic systems, especially with regard to human-robot collaboration in search and rescue (SaR) missions.  Often, people visiting public spaces like shopping malls, libraries, parks, etc. are unaware of their local map and exit points. Thus, in an emergency situation, a survivor can entrust a robot which has local map information to navigate out of the area. In turn, the robot can benefit from human's keen perception capability to ensure safety while navigating. To this end, we consider a modified version of the guide robot problem \cite{kanda2009affective}, adapted in the SaR situation where a robot helps a human in evacuating a hazardous environment.}

\begin{figure}
    \centering
    \includegraphics[trim= 0 20 50 20, clip, width=0.48\textwidth]{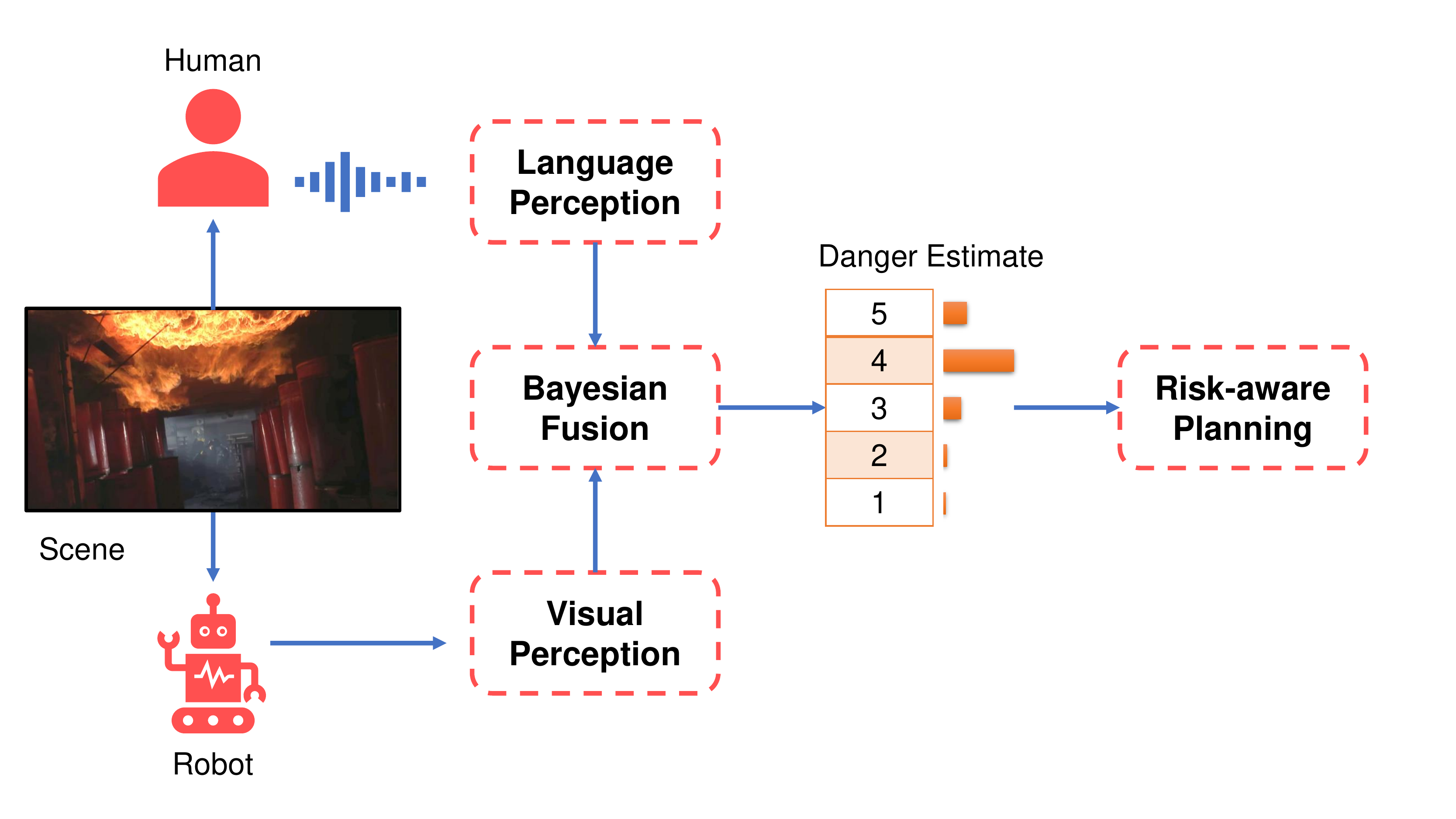}
    \caption{System Architecture. The human and the robot operating in the same environment perceive the world through their respective senses. Both the modalities are processed separately to extract meaningful information about potential danger, which gets then fused into a final estimate which is passed to the risk-aware planner.} 
    \label{fig:fullPipeline}
\end{figure}

The 2018 DARPA Subterranean challenge \cite{darpa2018sub} brought into light some of the key technical challenges that robots face when deployed in extreme environments \cite{agha2021nebula}.
Such environments possess severe obstruction for current perception systems due to low light conditions, sparse features, and presence of smoke, fire, fog, water, etc. 
Prior work has focused on extracting low-level features like occupancy grids \cite{agha2019confidence} or feature maps \cite{mur2015orb} from such complex environments. However, planning an evacuation requires the agent to make decisions based on high-level scene attributes like ``danger'', as pointed out in \cite{shree2021exploiting}. We thus propose a principled way to assess danger in a scene and leverage the danger information for planning a safe evacuation mission out of the hazardous environment. Furthermore, while there are merits to having a fully autonomous system, we believe that there is abundant opportunity to benefit from collaboration between a robot and the human agent, e.g. building trust. Thus, we propose a collaborative estimation strategy that takes advantage of human perception, contributing towards more successful evacuation from the environment.

There has been substantial improvement in visual perception performance in the past years, achieved by the application of deep convolutional neural networks  \cite{milz2018visual, socher2012convolutional, girshick2014rich, 
choy20163d}. The improvement comes at the cost of collecting large training data that can aptly resemble the test environment. Ironically, replicating hazards is tough due to local policies regarding safety and the high cost investment needed. These limitations have pushed researchers to test perception systems on simulated datasets \cite{jeon2019disc, wang2020tartanair, kirsanov2019discoman} or model environments \cite{kim2015real}. However, these alternatives are yet to attain the richness of the real world. To bridge this gap, we propose a scalable approach by leveraging images from movies which are significantly more photo-realistic than existing simulated datasets used in the community and can depict wider variety of scenarios as compared to the state-of-the-art model environments. \vikram{Moreover, this route avoids the privacy infringement concerns that using actual images from disaster sites can have.}

Our approach, sketched in Fig.~\ref{fig:fullPipeline}, initially consists of independent vision and language perception modules, which estimate the danger based on images from robot's sensor and verbal input from the user, respectively. The estimated danger levels are then fused together to get an improved danger estimate. 
Finally, taking advantage of the danger estimate, we propose a risk-aware planner that maximizes the chances of survival for the human-robot team.
In summary, this paper makes the following novel contributions:

\begin{enumerate}
    \item Development of a visuo-lingual dataset for perception in hazardous environments, with images taken from mass media that closely depict real world scenarios. The dataset entails distribution of danger as well as associated word descriptions of each scene.
    
    \item Performance assessment of representative machine learning models for danger estimation from images and language-input from humans.
    
    \item A Bayesian fusion framework that capitalizes on the likelihood models for both sensing modalities: vision and language, resulting in superior danger estimation.
    
    \item \vikram{Extensive testing with the collected dataset to evaluate our risk-aware mission planner, showing that the proposed approach enhances the success rate on average by $19\%$ points (compared to a baseline shortest path planner).}
    
\end{enumerate}

\section{Related Work}
\label{sec:lit_review}

\subsection{Cooperative rescue missions}
\label{subsec:lit_cooperative_missions}

In \cite{queralta2020collaborative}, the authors provide a comprehensive outline of the technical challenges encountered in the domain of multi-robot SaR. They remark that in order to fully benefit from the multi-robot team, there should be mechanisms to fuse information from different agents, enabling superior scene awareness. Our present work contributes along this direction with a collaborative scene perception pipeline in a human-robot team setup, where information is fused across the visual and the language domain.

The idea of collaborative decision making is inspired by humans. In an unforeseen disaster event, if a group of people are trapped together, they naturally tend to join hands to escape that situation. Consequently, researchers have tried to incorporate collaboration in multi-robot teams \cite{stroupe2005behavior} and in human-robot teams \cite{chandrasekaran2015human}. 
\vikram{Still}, prior studies suggest that human-robot teaming is relatively new in \vikram{rescue missions} because of interaction being a major bottleneck \cite{kruijff2014experience}. Our fusion framework allows the robot to account for human feedback, enabling superior danger awareness of the surrounding. \vikram{This knowledge about the environment can be further used by the robot for planning purposes, for instance to identify safer routes to an exit.}





\subsection{Disaster scene understanding}
\label{subsec:lit_scene_assement}
Safe navigation in SaR missions relies heavily upon accurate scene understanding. There are several tasks aiming at scene understanding like object recognition, semantic segmentation, physics-based reasoning, 3D reconstruction etc., as mentioned in \cite{naseer2018indoor}. While some of these tasks involve low-level reasoning like 3D reconstruction, others need high-level scene awareness like physics-based reasoning. In this work, we intend to leverage a high-level attribute of the environment, the notion of scene danger.

Often, danger is associated with the presence of fire or smoke in the scene, thus, prompting researchers to identify them in a scene \cite{li2020image,gaur2020video}. 
In contrast to these methods, we propose a more holistic danger perception of the environment which is not just limited to fire and smoke. Given a camera image from the environment, our perception module leverages state-of-the-art classification networks \cite{simonyan2014very, he2016deep, tan2019efficientnet} to predict a danger hypothesis. Furthermore, we embark on the opportunity to do collaborative perception for the task at hand and add a significant human component to get an updated danger representation of the environment.
\vikram{Previously, Ahmed et al. \cite{ahmed2012bayesian} introduced hybrid continuous-to-discrete likelihoods for fusing language data from humans by assuming a codebook consisting of a small set of words for the user to choose from.
%
In contrast, our model provides the freedom to the user to choose any word from the vast English vocabulary. Similar work has been pursued in \cite{shree2021exploiting} where the authors propose an adaptable danger estimation pipeline that relies on an a priori list of danger descriptions from an expert. There are two key aspects that differentiates our current work from \cite{shree2021exploiting}. First, our model does not necessitate language input from the human for danger estimation and is capable of assessing danger solely from camera data. This is advantageous in a scenario where the human is unable to provide feedback regarding their surrounding, e.g. due to cognitive impairment. Second, our multi-modal Bayesian framework allows multiple online updates based on incoming language data from the human, which is in contrast to \cite{shree2021exploiting} where the authors assume an a priori set of danger descriptions from an expert, specific to the environment.
}



\section{System Pipeline and Notation}
\label{sec:system}

\vikram{In our modified guide robot problem, we assume that the robot is present in the vicinity of a human survivor, who can follow the robot's path. The robot is capable of perceiving the environment through its camera sensor, and receives language input from the human about their surrounding. Furthermore, we assume that the robot has knowledge of the metric map of the environment and its objective is to find the \textit{best} path to an exit.}
The overall system can be divided into four major components as shown in Fig.~\ref{fig:fullPipeline}: visual perception, language perception, Bayesian fusion, and risk-aware planning. Following \cite{shree2021exploiting}, we assume a 5-point danger scale: 1-low, 2-moderate, 3-high, 4-very high, and 5-extreme. Let us denote an image from the environment by $I$ and its ground truth danger level by $D$. The visual perception module distills the key features of image $I$ and predicts an estimate for danger $y_{V}\in \{1,\cdots,5\}$. 

Assuming language inputs consisting of a single word from the human, let us denote the word input by $W$. The language module predicts a danger estimate $y_{L}\in\{1,\cdots,5\}$ based on the severity of word $W$. As a last step of the perception segment, the fusion module estimates a probability mass function (PMF) over the danger space, given the image and language input i.e. $\hat{D} = p(D=d|y_{V},y_{L})$, where $d\in \{1,\cdots,5\}$.

We assume that the robot knows the start and goal locations. The planner capitalizes on the danger estimate from the perception segment and plans an escape path that maximizes the \vikram{survival probability}. The following sections will elaborate each segment of our system in greater detail, starting first with our hazardous environment dataset.



\section{Hazardous Environment Dataset}
\label{sec:dataset}
To get authentic perceptual data that can replicate hazardous environments in the real world, we pool images from the vast collection of video clips that are easily accessible on various online platforms like Netflix, Xfinity, and Amazon Prime. The images are then annotated with the help of Amazon Mechanical Turk (AMT) workers and post-processed to generate ground truth labels. The dataset is provisioned under the fair use clause of copyrighted material and is opensourced for free usage by only the research community\footnote{Dataset available at \href{https://github.com/vikshree/hazard_dataset.git}{https://github.com/vikshree/hazard\_dataset.git}}.

\subsection{Image Selection}
\label{subsec:imageSelection}
We collect images from a wide range of movies and TV shows as candidates for our dataset. We initially select a set of 15 movies and 5 TV shows that embody scenes from variety of disaster scenarios, such as fire, flooding, and earthquakes. This is followed by capturing 75 small clips of the relevant sections in the movie/show with a screen capture software. Each clip (scene) is comprised of its unique set of visual and geographic  attributes, distinguishing it from the other scenes. Images are then extracted automatically from these scenes at 2 frames per second. At last, we perform a manual check to get rid of images that are either redundant or blurred, resulting in a total of 1002 images.

\subsection{Data annotation}
\label{subsec:annotation}
For annotation, we used Amazon Mechanical Turk (AMT). For each image, an AMT user must provide a danger rating from 1 to 5 and at most three keywords describing what led them to choose that specific danger rating. Thus, given an image $I$, we can represent the input from a AMT user $i$ as $(\eta_{i}, \mathbf{\Pi}_{i})$, where $\eta_i$ denotes the danger rating and $\mathbf{\Pi}_{i}$ is the set of keywords. To ensure data annotation quality, only AMT users with at least 1000 prior completed assignments and 98$\%$ approval rating were considered. Each image was examined by 15 unique AMT users, amounting to 355 unique AMT users.



\subsection{Data Statistics}
\label{subsec:stats}

\begin{figure}
	\centering
	\includegraphics[trim= 0 0 20 0, width=0.4\textwidth]{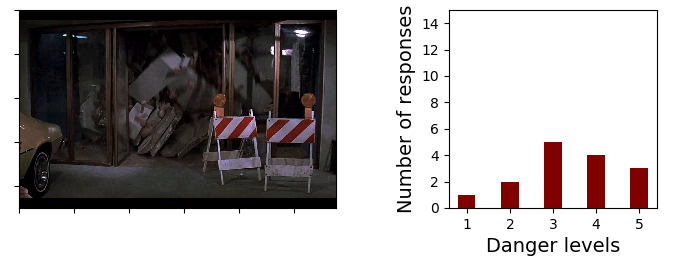}
	\includegraphics[trim= 0 0 5 0, width=0.4\textwidth]{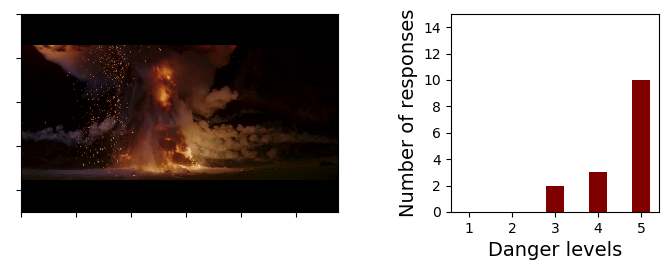}
	\caption{Images from our hazardous environment dataset and corresponding ratings by AMT users. Best viewed in color.}
	\label{fig:sampleImages}
\end{figure}

\begin{figure}
     \centering
     \begin{subfigure}[b]{0.24\textwidth}
         \centering
         \includegraphics[width=\textwidth]{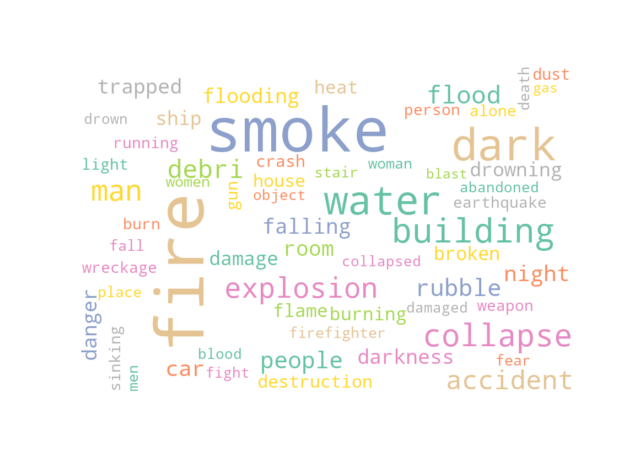}
         \caption{ }
         \label{fig:wordcloud}
     \end{subfigure}
     \hfill
     \begin{subfigure}[b]{0.23\textwidth}
         \centering
         \includegraphics[width=\textwidth]{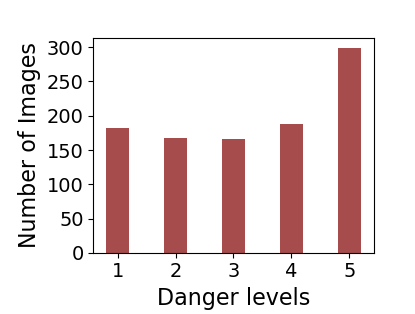}
         \caption{ }
         \label{fig:modedist}
     \end{subfigure}
        \caption{(a) Frequent words appearing in the responses of AMT users. (b) Distribution of mode danger rating of images across for the whole dataset. }
        \label{fig:qualitativeInsights}
\end{figure}

Our data collection results in 15K danger ratings and 45K associated keywords from the AMT users, with 3K unique words in it. A few sample images are shown in Fig.~\ref{fig:sampleImages}. We observe higher consensus in the danger ratings for images with extremely low or extremely high danger, as indicated in Fig.~\ref{fig:sampleImages}. To gain further insight into the language data, we show the wordcloud in Fig.~\ref{fig:wordcloud} depicting frequently used words. We find that factors like \textit{fire, smoke, water, dark,} and \textit{collapse}, play a key role in determining the danger rating.

Treating the \textit{mode} of the 15 danger responses for an image as representative of its ``true'' danger level, we show the count of images in our dataset belonging to each ``true'' danger category in Fig.~\ref{fig:modedist}. We observe that the number of images corresponding to different danger levels are comparable, thus, making the dataset well-balanced which is a crucial factor for training machine learning models.


\section{Visual Perception}
\label{sec:lang_danger}

\subsection{Task and Metrics}
\label{subsec:visTasks}
Given a local image $I$ obtained from the robot's camera sensor, the goal of the vision module is to distill its key aspects, and predict a danger rating for the image $y_{V} \in \{1,2,\cdots,5\}$. The model's ability to perceive entities that add to a person's notion of danger like fire, smoke, darkness, leakage, etc., is key to success in the task.

We use three metrics to capture the nuances in danger prediction.
First, it is typical to use top-1 accuracy for classification tasks and is defined as the proportion of time danger prediction $y_{V}$ matches the ``true'' danger of the image (assumed to be the  \textit{mode} $\tilde{D}$ of the AMT user-based danger ratings). Second, we use the root mean squared error of the predicted danger from the ``true'' danger of the image, i.e. 
\begin{equation}
    \text{RMSE} = \sqrt{\frac{1}{n} \sum_{k=1}^{n} (y_{V,k} - \tilde{D}_k)^2},
\label{eq:msd}
\end{equation}
where, $n$ is the total number of images. In addition to these standard metrics, we define a third one: the ``off-by-1'' accuracy. Off-by-1 accuracy is the proportion of time the danger estimate differs from $\tilde{D}$ at-most by 1 danger unit. 

\subsection{Performance Evaluation}
\label{subsec:visEval}
\subsubsection{Baselines}

We evaluate the ability of state-of-the-art (SOTA) models to estimate danger in hazardous environments. We select four candidate models: VGGNet \cite{simonyan2014very}, ResNet \cite{he2016deep}, 
DenseNet \cite{huang2017densely}, and EfficientNet \cite{tan2019efficientnet}. Each of these candidates have certain fundamental traits that differentiate them from one another. VGGNet is one of the oldest deep neural networks for image processing. ResNet addresses the vanishing gradients problem in deep networks by adding identity connections between layers. DenseNet uses dense blocks that receive features from its  preceding layers and also pass the processed features to all its subsequent layers, leading to stronger feature propagation. EfficientNet emphasizes scaling-up the network in a structured manner, leading to smaller yet effective models. The goal here is to assess performance and identify the most suitable model for our multi-modal danger assessment pipeline.


\subsubsection{Dataset}
\label{subsubsec:visDataset}
 We split our images and danger ratings data into train, validation, and test sets such that there are no overlapping scenes in two different sets. These sets were created manually, ensuring that the danger distribution for all three sets are similar. Ultimately, the train set consists of 795 images from 56 scenes, the validation set consists of 106 images from 10 scenes, and the test set consists of 101 images from 8 scenes.

\subsubsection{Training}
During training, the parameters for the last classification layer are tuned, while keeping the rest of the network frozen.
This takes advantage from the rich knowledge of the pre-trained model, \vikram{aligning well with our experiments where we observed higher performance across all metrics as compared to training the whole network}. 
Given an image $I$, its ground truth danger PMF, denoted by $\textbf{p} = [p_1, p_2, \cdots, p_5]$, is obtained by normalizing the corresponding ratings by the 15 users and \vikram{is used for training the models}. Each baseline network outputs a danger confidence $\textbf{c} = [c_1, c_2, \cdots, c_5]$. The vision-based danger estimate is defined as $y_{V} = \arg \max c_i \ \forall \ i \in \{1,2,\cdots,5\}$. While training we minimize KL divergence of the danger PMF $\textbf{p}$ from the model confidence $\textbf{c}$, i.e.
\begin{equation}
    D_{KL}(\textbf{p}||\textbf{c}) \equiv \sum_{i=1}^{5} p_{i} \big( \log p_{i}  - \log c_{i} \big).
\end{equation}

\subsubsection{Results}
\label{subsubsec:visionResults}
We train each baseline for 50 epochs and the best model is identified as the one with highest top-1 accuracy on the validation set. The performance of the best model for each baseline on the test set is reported in Table~\ref{tbl:visDangerResults}. All the models attain much higher top-1 accuracy compared to a randomized baseline, that would yield a top-1 accuracy of 20$\%$. For example, VGGNet-13 is correct about 48$\%$ of the times in predicting the right danger level for an image and about 80$\%$ of the time its estimate is at-most off-by-1 from the correct answer. Although the models are competitive, we chose VGGNet-13 for successive sections of the paper because of its best performance across all three metrics.

    

\begin{table}
\vspace{0.1in}
\centering
\caption{ \small Visual danger assessment performance of SOTA networks. Best performance is shown in \textbf{bold}. The number next to the architecture denotes a particular version.}
\begin{tabular}{|p{2cm}|p{1.2cm}|p{1.2cm}|p{1.2cm}|}
\hline
  \textbf{Model} & \textbf{Top-1} & \textbf{RMSE} & \textbf{Off-by-1}\\ 
  \hline 
  VGGNet-11 &  43.6 & 1.46 & 67.3\\  
  VGGNet-13 &  \textbf{47.5} & \textbf{1.24} & \textbf{79.2}\\
VGGNet-16 &  46.5 & 1.52 & 71.3\\
   \hline
  ResNet-50 & 39.6 & 1.55 & 68.3 \\
  ResNet-101 & 45.5 & 1.40 & 75.2 \\
  ResNet-152 & 42.6 & 1.59 & 70.3 \\
 \hline
  DenseNet-121 & 30.7 & 1.78 & 63.4 \\
  DenseNet-169 & 38.6 & 1.73 & 65.3 \\
  DenseNet-201 & 32.7 & 1.76 & 63.4 \\
 \hline
  EfficientNet-b0 & 44.6 & 1.45 & 72.3 \\
  EfficientNet-b1 & 31.7 & 1.85 & 61.4 \\
  EfficientNet-b2 & 34.7 & 1.56 & 68.3 \\
  EfficientNet-b3 & 36.6 & 1.58 & 72.3 \\
  EfficientNet-b4 & \textbf{47.5} & 1.49 & 72.3 \\
  EfficientNet-b5 & 34.7 & 1.70 & 63.4 \\ 
 \hline
  Randomized & 20.0 & 2.0 & 52.0 \\ 
 \hline
\end{tabular}\\
\label{tbl:visDangerResults}
\end{table}

\begin{figure}
     \centering
     \begin{subfigure}[b]{0.45\textwidth}
    \centering
	\includegraphics[trim= 50 410 400 0, width=0.85\textwidth,height=50pt]{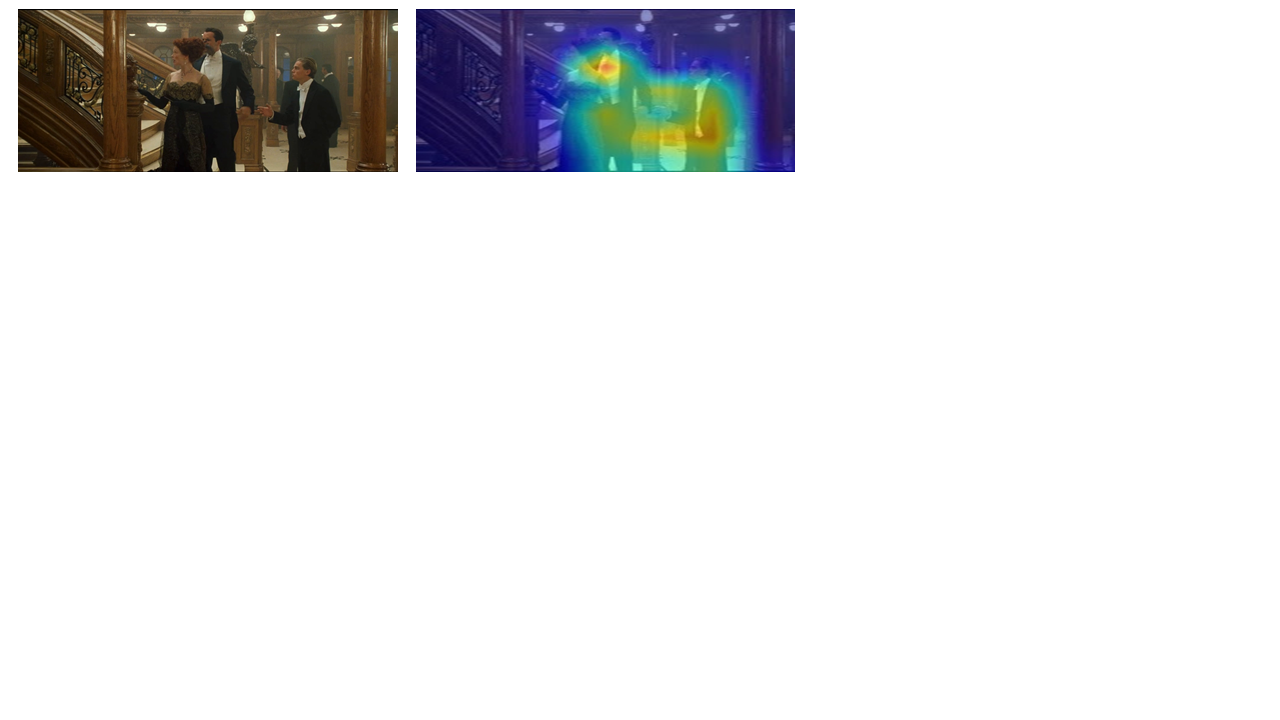}
    \caption{Output Danger = 1 }
    \label{fig:danger1}
     \end{subfigure}
     \begin{subfigure}[b]{0.45\textwidth}
     \centering
	\includegraphics[trim= 50 370 400 0, width=0.85\textwidth,height=50pt]{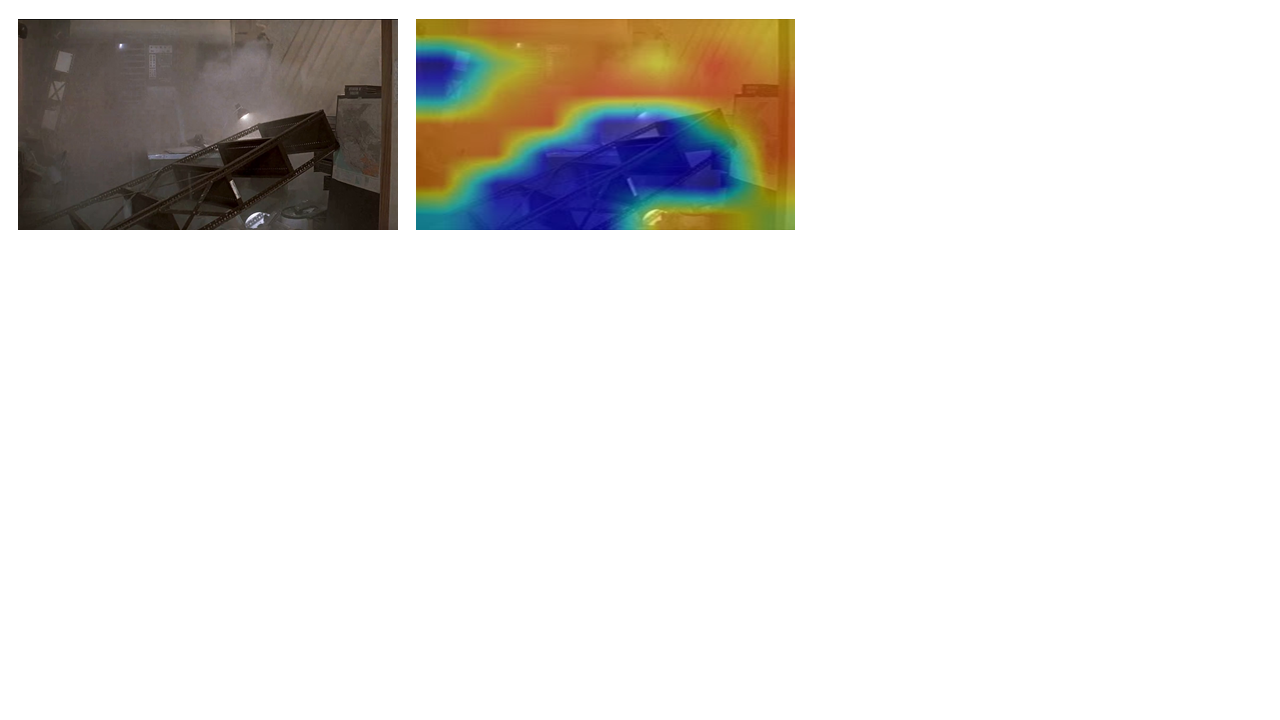}
    \caption{Output Danger = 3}
    \label{fig:danger3}
     \end{subfigure}
     \begin{subfigure}[b]{0.45\textwidth}
     \centering
	\includegraphics[trim= 50 410 400 0, width=0.85\textwidth,height=50pt]{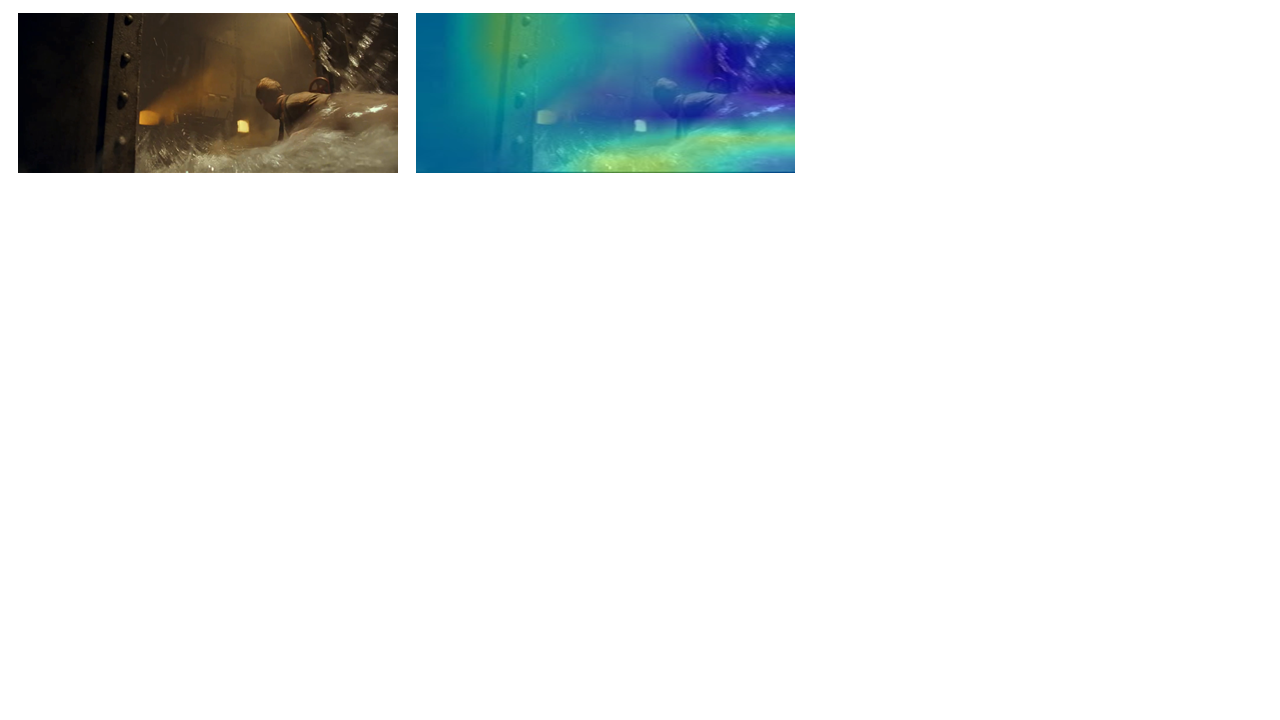}
    \caption{Output Danger = 5}
    \label{fig:danger5}
     \end{subfigure}
	\caption{Grad-CAM visualizations for test set images with VGGNet-13 model predictions. \vikram{Red regions provide the highest contribution while blue regions the lowest. }Best viewed in color.}
	\label{fig:gradCamPics}
\end{figure}

For intuitive understanding of the model predictions, we leverage the Gradient-weighted Class Activation Mapping \cite{selvaraju2017grad} (Grad-CAM) visualization tool. Grad-CAM produces a color map highlighting the important regions of the image that contributed towards the predicted result. This not only provides the much needed insight into the deep learning model, \vikram{but could also help gain the trust of human when used in real-world missions.} For example, we can observe that  the image in Fig.~\ref{fig:gradCamPics} showing some people chatting has a low predicted danger.
The smoke in Fig.~\ref{fig:gradCamPics}b is a significant contributor for its danger prediction of 3. Finally, the flooded area in Fig.~\ref{fig:gradCamPics}c prompts the model to predict extreme danger. The quantitative as well as the qualitative results demonstrate the ability of our baseline models to learn and predict danger from images. These baseline results are a solid precursor to specialized models for vision-based danger assessment, which is part of our future work.


\section{Language Perception}
\label{sec:lang_danger}
\subsection{Task and Metrics}
\label{subsec:langTasks}
The language perception module takes into account the word input $W$ from the human and accordingly predicts a danger rating $y_L \in \{1,2,\cdots,5\}$. The ability to understand what words people tend to use in hazardous environmental conditions is key to success in this task. It is intuitive to use the same metrics defined in section~\ref{subsec:visTasks} for assessing performance of language perception models i.e. top-1 accuracy, root mean squared error (RMSE), and off-by-1 accuracy.

\subsection{Performance Evaluation}

\subsubsection{Baselines}
\begin{figure}
    \centering
    \includegraphics[trim= 10 310 10 55, clip, width=0.46\textwidth]{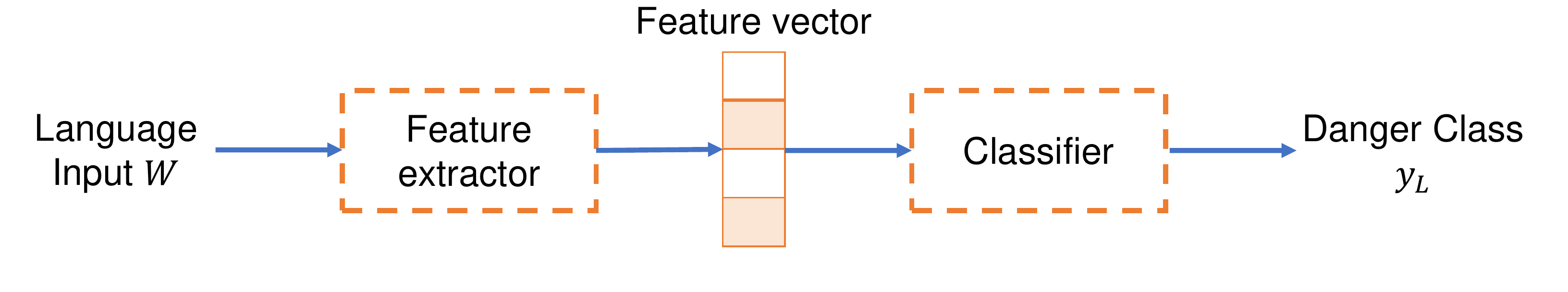}
    \caption{Language perception pipeline. First, words are converted into features, followed by a classifier that predicts danger $y_{L}$.}
    \label{fig:languagePipeline}
\end{figure}

\vikram{Following the mainstream approach} in language classification, we first convert the words into a low-dimensional feature vector, which is then passed to a classifier \cite{goldberg2017neural}. This is shown in Fig.~\ref{fig:languagePipeline}. In this work, we use GloVe features \cite{pennington2014glove} for their superior performance reported in literature as compared to other representations like Word2Vec \cite{mikolov2013efficient}. We test danger prediction with three well-known candidate classifiers: K nearest neighbor (KNN), logistic regression, and Support Vector Machine (SVM). The KNN algorithm predicts the danger class based on the k nearest matches of the word $W$ in the training data; logistic regression aims at maximizing the conditional likelihood of the training data;
SVM maximizes the margin between class variables, making it less prone to outliers compared to logistic regression. 

\subsubsection{Dataset}
\label{subsubsec:langDataset}
For evaluating performance of danger prediction based on language input, we use the keywords associated with the images and their corresponding danger ratings from the hazardous environment dataset. Since about $90\%$ of the keywords consist of a single word, we assume one-word input from the human and ignore the sentences in our dataset. A typical training data point is of the form $(W,\eta)$, where $W$ is the word describing the scene and $\eta$ is its danger level. We use the same splits for training, validation, and test set as chosen in the visual perception case, yielding a total of 31K, 4K, and 4K keywords in these sets, respectively.



\subsubsection{Results}
The results based on the test set are shown in Table~\ref{tbl:langDangerResults}. SVM's compatibility with high-dimensional data renders it superior top-1 accuracy compared to other models. All three class of models achieve competitive performance in terms of RMSE and off-by-1 accuracy.
It is interesting to note that the values of \vikram{all the} metrics for language-based danger predictor are significantly lower than the best visual perception model. This can be attributed to the richness of visual data when compared to single word inputs from the human.
Since top-1 accuracy is widely accepted in the literature, we chose to use SVM (RBF kernel) for the subsequent sections.

\vikram{Table~\ref{tbl:qualitativeLanguageResults} shows a few words frequently appearing in the test set along with their corresponding danger prediction by the model. Words with danger prediction of 1, such as people, gathering, dirty etc., align well with our intuitive sense of danger. Similarly, words that have a danger prediction of 5, such as fire, explosion, flooding etc, are also straightforward. Note that words with intermediate danger, such as debris, broken, crash etc. are in fact contentious for humans because of the lack of sufficient information that these words bear.}


\begin{table}
\centering
\caption{ \small Language perception performance. Best performance is shown in \textbf{bold}.}
\begin{tabular}{|p{2.5cm}|p{1cm}|p{1cm}|p{1.2cm}|}
\hline
  \textbf{Model} & \textbf{Top-1} & \textbf{RMSE} & \textbf{Off-by-1}\\
  \hline 
  1-NN & 27.2 & 1.72 & 64.5\\
  3-NN & 28.8 & 1.79 & 61.0\\
  5-NN & 30.8 & 1.70 & 66.3\\
  11-NN & 32.4 & \textbf{1.58} & 68.4\\
   \hline
  Logistic Regression & 36.9 & 1.67 & 67.1 \\
 \hline
  SVM (Linear) & 37.1 & 1.68 & 66.8 \\
  SVM (Poly. kernel) & 37.1 & 1.64 & \textbf{68.5} \\
  SVM (RBF kernel) & \textbf{37.6} & 1.63 & \textbf{68.5} \\
 \hline
   Randomized & 20.0 & 2.0 & 52.0 \\ 
 \hline
\end{tabular}\\
\label{tbl:langDangerResults}
\end{table}

\begin{table}
\centering
\caption{Common words and their danger predictions generated from the SVM model.}
\begin{tabular}{|p{5.5cm}|p{2cm}|}
\hline
  \textbf{Words} & \textbf{Danger output} \\ 
  \hline 
  people, gathering, dirty,  night, tunnel &  1 \\
  \hline
  darkness, sewer, dust, cave, broken &  2 \\
  \hline
  debris, suffocation, wreckage, damage, violence &  3 \\
  \hline
  flood, accident, weapon, crash, freezing &  4 \\
  \hline
  fire, explosion, collapse, flooding, earthquake   &  5\\
 \hline
\end{tabular}
\label{tbl:qualitativeLanguageResults}
\end{table}
\section{Fused Danger Assessment}
\label{sec:fusion}

Vision and language perception modalities both have their own benefits. On one hand, image data is richer compared to words. On the other hand, language modality can take advantage of human's keen perception to focus on key entities that contribute towards danger in the scene.
Thus, it is natural to combine both modalities with the goal to achieve superior danger perception. 
Furthermore, as pointed out in the literature \cite{eder2014towards}, accounting for human feedback enables a human-in-the-loop mission approach, thus promoting trust between the human and the robot.
Hence, we now introduce our Bayesian fusion framework.


\subsection{Bayesian fusion}
\label{subsec:bayesTheory}
The goal of the Bayesian fusion module is estimate $\hat{D} \equiv p(D=d|y_{V},y_{L})$, where $d\in \{1,\cdots,5\}$. From Bayes rule, assuming conditional independence of predictions from the visual and language modules, given the scene danger $d$:
\begin{align}
\nonumber
p(D=d|y_{V},y_{L}) & \propto p(y_{V}, y_{L}|D=d)p(D=d)\\
& \propto \ p(y_{V}|D=d) p(y_{L}|D=d) p(D=d),
\label{eq:fusion}
\end{align}

\noindent where $p(y_{V}|D=d)$ and $p(y_{L}|D=d)$ denote the vision-based and language-based sensing likelihoods, respectively. Eq.~\eqref{eq:fusion} is key for our fusion module. 

Note that until now, we have assumed a single word input from the human. However, this assumption can now be relaxed and Eq.~\eqref{eq:fusion} can be extended to $m$ number of human inputs:
\begin{align}
    \nonumber
    &\hat{D} \equiv p(D=d|y_{V},y^{1}_{L},\cdots,y^{m}_{L})\\
    \Rightarrow &\hat{D} \propto p(y_{V}|D=d) \bigg( \prod_{k=1}^{m} p(y^{k}_{L}|D=d) \bigg) p(D=d).
    \label{eq:multipleFusion}
\end{align}
As evident from Eq.~\ref{eq:multipleFusion}, we need the likelihood functions for visual and language perception models. First, consider the vision-based danger likelihood function and denote it by $l_{V}^{i,j} \equiv p(y_{V}=i|D=j)$, where $i, j \in \{1,\cdots,5\}$. The likelihood function can be calculated from the validation set:
\begin{align}
\nonumber
l_{V}^{i,j} & \equiv p(y_{V}=i|D=j) \\
& \approx \frac{\text{\# images with ``true'' danger } j \text{ and prediction } i}{\text{\# images with ``true'' danger level }j}.
\label{eq:visLikelihood}
\end{align}
In the lack of a large validation set, as in our case, Eq.~\eqref{eq:visLikelihood} may overfit to the set and cause poor performance on the test set. To avoid this problem, we apply $K$-fold cross-validation strategy where  depending on the selection of the validation set, we get $K$ different estimates for the likelihood function $l_{V}^{i,j}$. 
Thereafter, the \textit{mean} likelihood function is obtained, i.e. $\hat{l}_{V}^{i,j} = \frac{1}{K}\sum_{k=1}^{K} l_{V, k}^{i,j}$.
We follow the same strategy to obtain the \textit{mean} language-based danger likelihood function $\hat{l}_{L}^{i,j}$.





\begin{figure}
    \centering
    \includegraphics[width=0.22\textwidth]{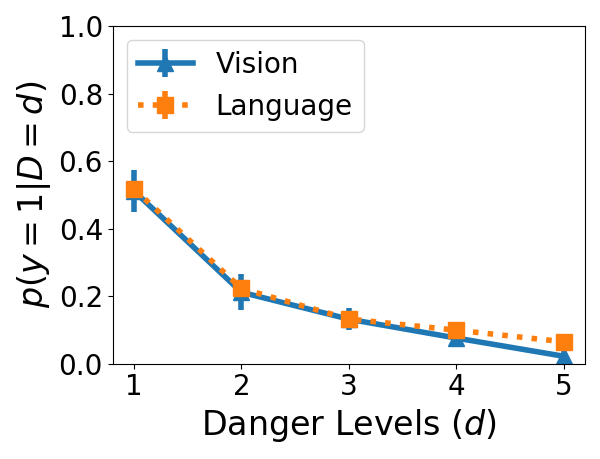}
    \includegraphics[width=0.22\textwidth]{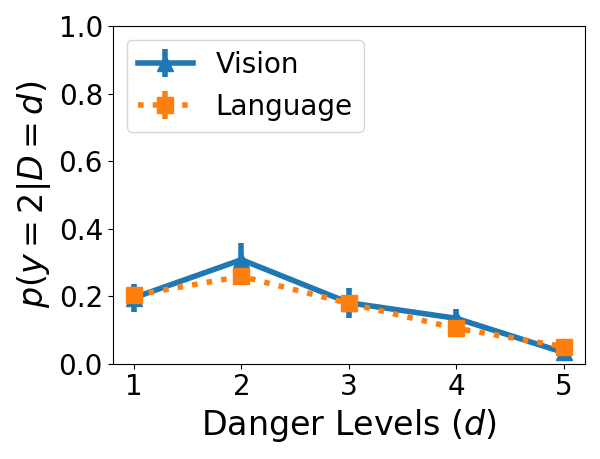}
    \includegraphics[width=0.22\textwidth]{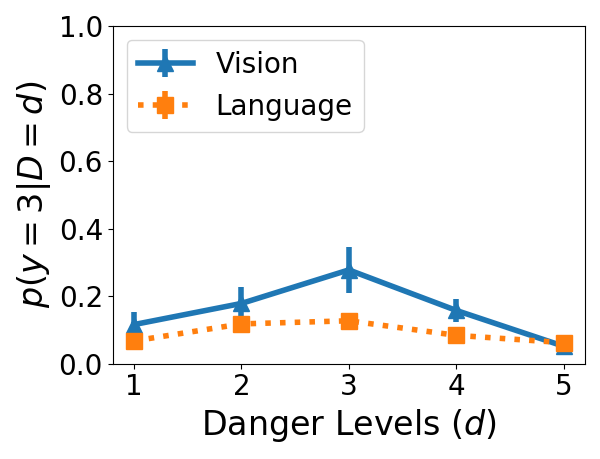} \includegraphics[width=0.22\textwidth]{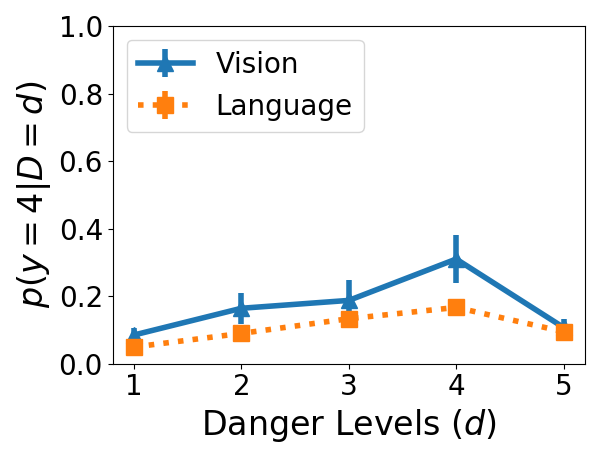} \includegraphics[width=0.22\textwidth]{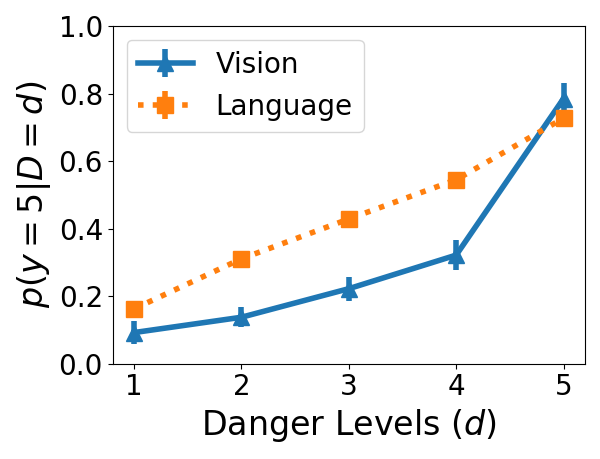}  
    \caption{Vision-based and Language-based likelihood function estimated from 9-fold validation.} 
    \label{fig:likelihoodVisLang}
\end{figure}
\subsection{Performance evaluation}
We apply 9-fold cross-validation to calculate the \textit{mean} likelihood functions $\hat{l}_{V}^{i,j}$ and $\hat{l}_{L}^{i,j}$. See Fig.~\ref{fig:likelihoodVisLang}.
Two crucial observations can be made here. First, we found that the maximum of the $mean$ likelihood function occurs at the ``true'' danger level, i.e. $\arg \max_{j} \hat{l}^{i,j}_{V} = \arg \max_{j} \hat{l}^{i,j}_{L} = i$. 
Second, the values of likelihood functions at the ``true'' danger level, i.e. $\hat{l}^{i,i}_{V}$ and $\hat{l}^{i,i}_{L}$, are higher at the two extremes i.e. $d=1$ and $5$. This is because both images and words with danger values at the extremes are easier to classify by our models as compared to the ones that belong to moderate danger.


To evaluate the performance of the fused danger estimate, we use the hazardous environment dataset with the same test set as chosen in section~\ref{subsubsec:visDataset}. \vikram{This helps us to gauge the relative change in performance by fusing the two modalities.} Given an image $I$ and corresponding set of keywords $\mathbf{\Pi}$, we simulate input from the human by randomly sampling keywords from $\mathbf{\Pi}$ without replacement. \vikram{To evaluate the estimation performance, we use the maximum a posteriori (MAP) estimate $\hat{D}_{\text{MAP}}=~\arg \max_{d} \ p(D=d|y_{V}, y_{L}^{1},\cdots,y_{L}^{m})$. Comparing the MAP estimate $\hat{D}_{\text{MAP}}$ with the ``true'' danger $\tilde{D}$,} we were able to evaluate the top-1 accuracy, off-by-1 accuracy, and RMSE with different number of word inputs from the human. For visual perception, we use the VGG-13 model, while for language perception we use the SVM (with RBF kernel) model. The results, presented in Table~\ref{tbl:multiModalDangerResults},
reveal that all the three metrics are maximized by leveraging the multi-modal pipeline. Specifically, combining visual data with 5 words from the human improves top-1 accuracy by $1\%$ point and RMSE by $17\%$, compared to using any of the single modalities. 
Note that although the Bayesian model is tested with 5 words, it can incorporate even higher number of word inputs from human.
However, special care should be taken to avoid redundant information because it can violate the independence assumption used in Eq.~\eqref{eq:multipleFusion}, leading the Bayesian model to overtrust the data coming from the human.




\begin{table}
\centering
\caption{Multi-modal danger assessment performance. Best performance is shown in \textbf{bold}.}
\begin{tabular}{|p{2.5cm}|p{1.2cm}|p{1.2cm}|p{1.2cm}|}
\hline
  \textbf{Modality} & \textbf{Top-1} & \textbf{RMSE} & \textbf{Off-by-1}\\ 
  \hline 
  Vision only &  47.5 & 1.24 & 79.2\\
  \hline
  Language-only & 37.6 & 1.63 & 68.5\\
 \hline
  VL 1-word & 46.5 & 1.30 & 78.2 \\
  VL 2-words & 47.5 & 1.13 & \textbf{84.2} \\
  VL 3-words &  44.6 & 1.27 & 80.2  \\
  VL 4-words & 47.5 & 1.11 & 83.2 \\
  VL 5-words & \textbf{48.5} & \textbf{1.03} & 83.2 \\
 \hline
\end{tabular}\\
\label{tbl:multiModalDangerResults}
\vspace{-0.1in}
\end{table}

\section{Risk-aware Planning}
\label{sec:planning}
With the capability to estimate danger, we now introduce the escape route planning problem.

\subsection{Planning Problem and Solution Approach}
\label{subsec:planningTasks}
Let us represent the environment as a directed graph $G=(V,A)$, where vertices $V$ represent a set of locations in the environment, and arcs $A$ represent the possibility to travel between two locations.
The starting vertex of the human-robot team is denoted by $v_{\text{s}}\in V$ and the goal location (for example, one of the building exits) is denoted by $v_{\text{g}} \in V$. 
Let us define a parameter $\tau~\vikram{\in \{1, ..., 5\}}$, denoting the level of danger that the human-robot team can tolerate. 
Accordingly, we define the survival probability of traveling along arc $(i,j) \in A$ as $s_{ij} = p(D_{j}\leq\tau)$, where $D_{j}$ is the ground truth danger of the destination node $j$. 
Assuming these events to be independent, the survival probability $s_{\pi}$ along a graph path $\pi=[(v_\text{s}=v_1, v_2),\ldots,(v_{k-1}, v_k=v_\text{g})]$ connecting start and goal vertices can be expressed as
\begin{align}
    s_{\pi} = \prod_{i=1}^{k}  s_{v_i v_{i+1}}.
    \label{eq:netSirvival}
\end{align} 
Note that the independence assumption might not always hold in practice, and more sophisticated models could be built to account for spatial dependencies in the danger map.
If all the survival probabilities were known exactly and in advance, the path $\pi^{*}$ maximizing the the overall probability of survival, i.e.
\begin{align}
    \pi^{*}~\vikram{\in}~\arg \max_{\pi} \ s_{\pi},
    \label{eq:optimizeCost}
\end{align}
\noindent could be easily obtained by computing the shortest path between $v_s$ and $v_g$ on a weighted version of the graph $G$, with weights $w_{ij}$ computed as $w_{ij}=-\log s_{ij}$. However, the robot does not have access to the true survival probabilities $s_{ij}$, and it must rely on the danger estimate $\hat{D}$ to establish an approximation for the survival probability $\hat{s}_{ij} = p(\hat{D}\leq\tau)$. We assume that the robot-human team can only access data of the neighboring vertices of $G$. Furthermore, the unexplored vertices are assigned a uniform prior danger distribution.

Given the problem inputs described above, the goal of the planning module is to compute a policy that maximizes the mission success rate, i.e. the probability of reaching $v_g$ without traversing an arc having a ground truth danger level higher than $\tau$. Since the survival probability in Eq.~\eqref{eq:optimizeCost} depends on robot's belief of danger, we hypothesize that superior danger perception with multi-modal sensing can enable the team to avoid hazardous exposures, leading to higher mission success. 

To tackle the planning problem described above, we use the following receding-horizon planning heuristic: at each planning iteration, the team moves along the first arc of the safest path $\pi^{*}$ computed as described above, but replacing $s_{ij}$ with $\hat{s}_{ij}$. When the corresponding destination vertex is reached, the team updates the danger estimate of the neighboring vertices. This process is repeated until the team ends up in a vertex with an intolerable danger level (in which case the mission counts as a failure), or the the goal node $v_{g}$ is reached.

\begin{figure}
    \centering \includegraphics[trim= 140 100 110 70, clip, width=0.38\textwidth]{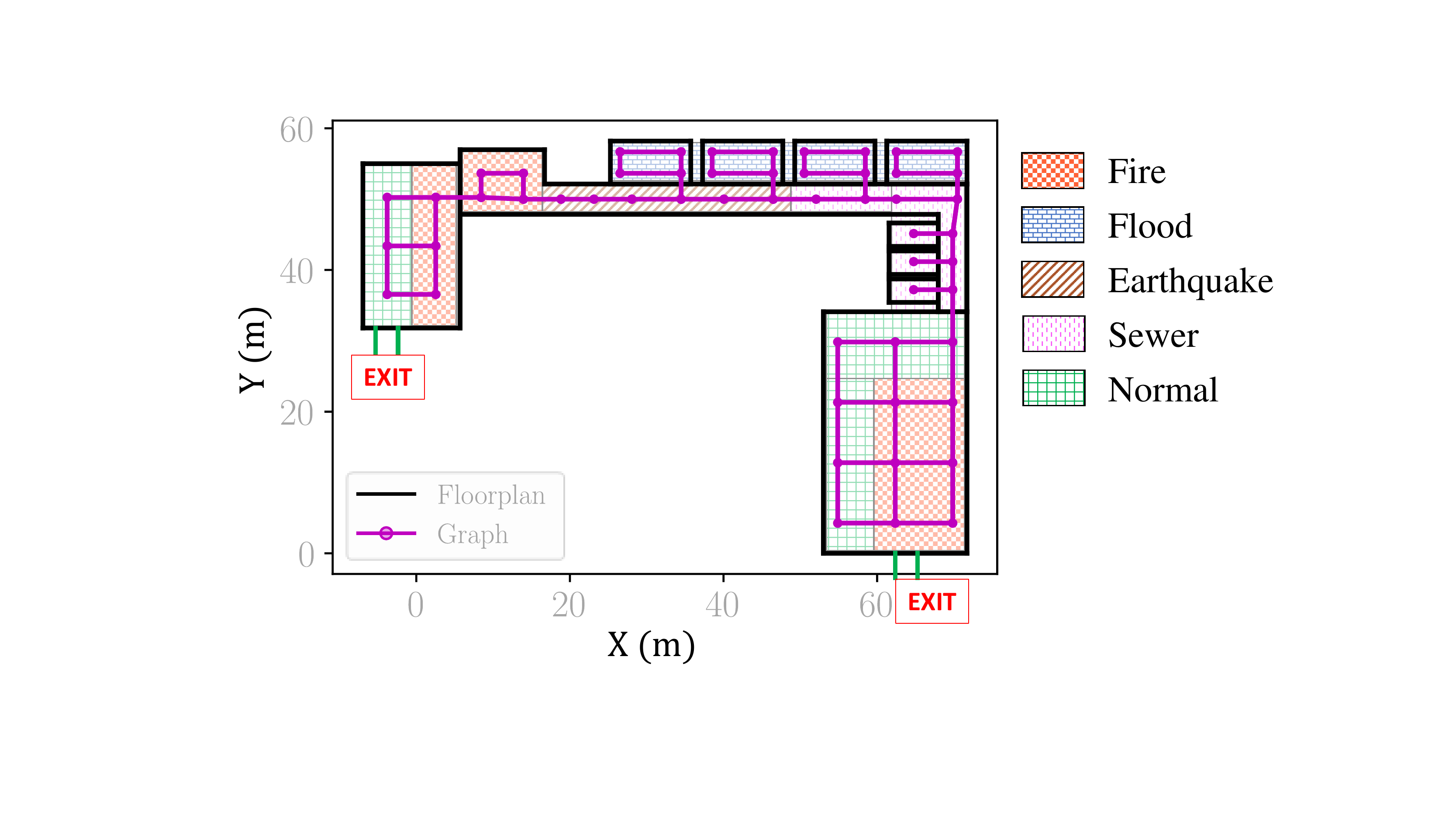}
    \caption{Graphical representation of the environment used in our simulations, with danger map.}
    \label{fig:schoolGraph}
    \vspace{-0.1in}
\end{figure}

\subsection{Simulation Enironment}
\label{subsubsec:simEnvironment}
We use the School environment from \cite{shree2021exploiting} and abstract it into the final graph shown in  Fig.~\ref{fig:schoolGraph}, consisting of a total of $n=54$ nodes and two exits. We manually assign scene characteristics e.g. fire, flood, etc., for different segments of the environment and accordingly associate each node with an image from the test set of the hazardous environment dataset.



\begin{figure}[!ht]
     \centering
    \includegraphics[trim= 0 0 0 0, clip, width=0.4\textwidth]{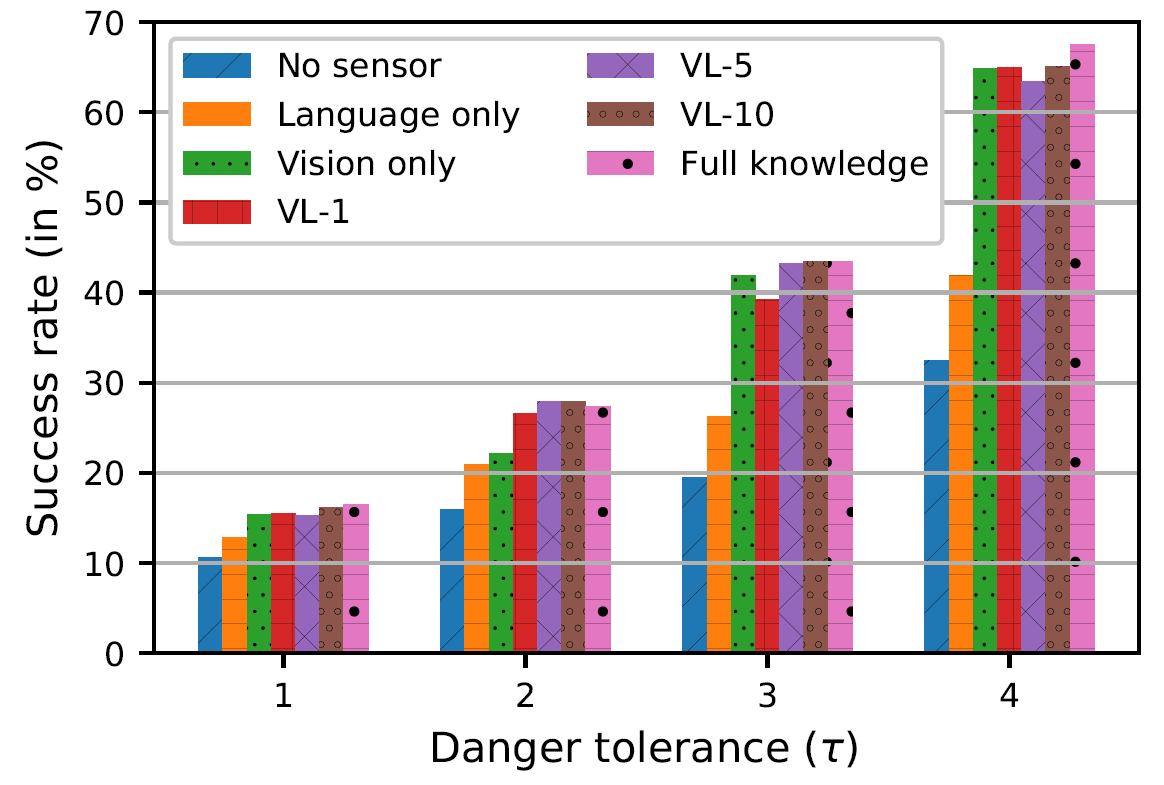}
        \caption{Planning results: Success rate with different sensing modalities. Results are obtained from 1000 simulation runs.}
        \label{fig:successRate}
        \vspace{-0.1in}
\end{figure}

\subsection{Results and Discussion}
We perform 1000 Monte Carlo simulations of the combined planning and danger estimation framework with different sensor modalities and for different tolerable danger levels $\tau$. The ``no-sensor'' case is when the robot is unable to perceive danger. The ``full-knowledge'' case refers to the hypothetical situation when the robot is aware of the ground truth danger map of the whole environment. 
During simulations, when the team is moving from node $i$ to $j$, the survival of the team is sampled from PMF of $s_{ij}$. A successful mission is the one where the team survives the whole mission and reaches the exit.
Note that we do not report results for $\tau=5$ since it corresponds to the unrealistic case when the team can survive extreme danger.   

In Fig.~\ref{fig:successRate}, we observe that the ``no sensor'' case has lowest success rate amongst all methods for every value of $\tau$. This indicates that the ability to perceive and account for danger is critical for mission success in disaster scenarios. A further analysis of different modalities provides some intriguing insights. The ``vision only'' sensing consistently outperforms ``language only'' sensing. As mentioned earlier, this is because of the richness of visual data compared to single word inputs. Occasionally, the ``vision-only'' success rate is even competitive to the multi-modal method, implying that the perception advantage shown in Table~\ref{tbl:multiModalDangerResults} does not always yield similar benefits in terms of mission success. 
Nonetheless, across all simulation scenarios, the highest success rate is achieved in VL-10 case, by fusing information between the vision and language domain. The results support our hypothesis that improved perception leads to more successful missions.
In fact, sometimes VL-10 case even outperforms the model with full knowledge. This is probably because of higher danger estimate reported by the model, compared to ground truth danger, thus, forcing the robot to choose a conservative route.
\section{Conclusion}
\label{sec:conclusion}

Our work demonstrates that leveraging the vast collection of visual content from mass media can enable perception systems to function in disaster scenarios. The hazardous environment dataset paves the way for development and testing of future danger assessment pipelines. Further, we show through simulations that compared to an autonomous robot that only relies on a single sensing modality, a collaborative robot that takes into account the feedback from human user is better equipped to estimate danger. Finally, our risk-aware planning framework translates the improvements in danger assessment into tangible metrics such as higher mission success rate, which is critical in search and rescue operations.


\bibliographystyle{IEEEtran} 
\bibliography{egbib} 

\begin{thebibliography}{10}
\providecommand{\url}[1]{#1}
\csname url@rmstyle\endcsname
\providecommand{\newblock}{\relax}
\providecommand{\bibinfo}[2]{#2}
\providecommand\BIBentrySTDinterwordspacing{\spaceskip=0pt\relax}
\providecommand\BIBentryALTinterwordstretchfactor{4}
\providecommand\BIBentryALTinterwordspacing{\spaceskip=\fontdimen2\font plus
\BIBentryALTinterwordstretchfactor\fontdimen3\font minus
  \fontdimen4\font\relax}
\providecommand\BIBforeignlanguage[2]{{%
\expandafter\ifx\csname l@#1\endcsname\relax
\typeout{** WARNING: IEEEtran.bst: No hyphenation pattern has been}%
\typeout{** loaded for the language `#1'. Using the pattern for}%
\typeout{** the default language instead.}%
\else
\language=\csname l@#1\endcsname
\fi
#2}}

\bibitem{casper2003human}
J.~Casper and R.~R. Murphy, ``Human-robot interactions during the
  robot-assisted urban search and rescue response at the world trade center,''
  \emph{IEEE Transactions on Systems, Man, and Cybernetics, Part B
  (Cybernetics)}, vol.~33, no.~3, pp. 367--385, 2003.

\bibitem{murphy2021how}
\BIBentryALTinterwordspacing
R.~R. Murphy, ``How robots helped out after the surfside condo collapse,''
  \emph{IEEE Spectrum}, 2021. [Online]. Available:
  \url{https://spectrum.ieee.org/building-collapse-surfside-robots}
\BIBentrySTDinterwordspacing

\bibitem{kanda2009affective}
T.~Kanda, M.~Shiomi, Z.~Miyashita, H.~Ishiguro, and N.~Hagita, ``An affective
  guide robot in a shopping mall,'' in \emph{HRI}, 2009, pp. 173--180.

\bibitem{darpa2018sub}
\BIBentryALTinterwordspacing
DARPA, ``{\MakeUppercase DARPA} subterranean challenge,'' 2018. [Online].
  Available: \url{https://www.subtchallenge.com}
\BIBentrySTDinterwordspacing

\bibitem{agha2021nebula}
A.~Agha \emph{et~al.}, ``Nebula: Quest for robotic autonomy in challenging
  environments; team costar at the darpa subterranean challenge,''
  \emph{Journal of Field Robotics}, 2021.

\bibitem{agha2019confidence}
A.-A. Agha-Mohammadi, E.~Heiden, K.~Hausman, and G.~Sukhatme, ``Confidence-rich
  grid mapping,'' \emph{IJRR}, vol.~38, no. 12-13, pp. 1352--1374, 2019.

\bibitem{mur2015orb}
R.~Mur-Artal, J.~M.~M. Montiel, and J.~D. Tardos, ``Orb-slam: a versatile and
  accurate monocular slam system,'' \emph{IEEE Transactions on Robotics},
  vol.~31, no.~5, pp. 1147--1163, 2015.

\bibitem{shree2021exploiting}
V.~Shree, B.~Asfora, R.~Zheng, S.~Hong, J.~Banfi, and M.~Campbell, ``Exploiting
  natural language for efficient risk-aware multi-robot sar planning,''
  \emph{RA-L}, vol.~6, no.~2, pp. 3152--3159, 2021.

\bibitem{milz2018visual}
S.~Milz, G.~Arbeiter, C.~Witt, B.~Abdallah, and S.~Yogamani, ``Visual slam for
  automated driving: Exploring the applications of deep learning,'' in
  \emph{CVPR Workshops}, 2018, pp. 247--257.

\bibitem{socher2012convolutional}
R.~Socher, B.~Huval, B.~Bath, C.~D. Manning, and A.~Ng,
  ``Convolutional-recursive deep learning for 3d object classification,''
  \emph{Advances in Neural Information Processing Systems}, vol.~25, pp.
  656--664, 2012.

\bibitem{girshick2014rich}
R.~Girshick, J.~Donahue, T.~Darrell, and J.~Malik, ``Rich feature hierarchies
  for accurate object detection and semantic segmentation,'' in \emph{CVPR},
  2014, pp. 580--587.

\bibitem{choy20163d}
C.~B. Choy, D.~Xu, J.~Gwak, K.~Chen, and S.~Savarese, ``3d-r2n2: A unified
  approach for single and multi-view 3d object reconstruction,'' \emph{ECCV},
  pp. 628--644, 2016.

\bibitem{jeon2019disc}
H.-G. Jeon, S.~Im, B.-U. Lee, D.-G. Choi, M.~Hebert, and I.~S. Kweon, ``Disc: A
  large-scale virtual dataset for simulating disaster scenarios,'' \emph{IROS},
  pp. 187--194, 2019.

\bibitem{wang2020tartanair}
W.~Wang, D.~Zhu, X.~Wang, Y.~Hu, Y.~Qiu, C.~Wang, Y.~Hu, A.~Kapoor, and
  S.~Scherer, ``Tartanair: A dataset to push the limits of visual slam,''
  \emph{IROS}, pp. 4909--4916, 2020.

\bibitem{kirsanov2019discoman}
P.~Kirsanov \emph{et~al.}, ``Discoman: Dataset of indoor scenes for odometry,
  mapping and navigation,'' \emph{IROS}, pp. 2470--2477, 2019.

\bibitem{kim2015real}
J.-H. Kim and B.~Y. Lattimer, ``Real-time probabilistic classification of fire
  and smoke using thermal imagery for intelligent firefighting robot,''
  \emph{Fire Safety Journal}, vol.~72, pp. 40--49, 2015.

\bibitem{queralta2020collaborative}
J.~P. Queralta \emph{et~al.}, ``Collaborative multi-robot search and rescue:
  Planning, coordination, perception, and active vision,'' \emph{IEEE Access},
  vol.~8, pp. 191\,617--191\,643, 2020.

\bibitem{stroupe2005behavior}
A.~Stroupe, T.~Huntsberger, A.~Okon, H.~Aghazarian, and M.~Robinson,
  ``Behavior-based multi-robot collaboration for autonomous construction
  tasks,'' \emph{IROS}, pp. 1495--1500, 2005.

\bibitem{chandrasekaran2015human}
B.~Chandrasekaran and J.~M. Conrad, ``Human-robot collaboration: A survey,''
  \emph{SoutheastCon 2015}, pp. 1--8, 2015.

\bibitem{kruijff2014experience}
G.-J.~M. Kruijff \emph{et~al.}, ``Experience in system design for human-robot
  teaming in urban search and rescue,'' in \emph{Field and Service
  Robotics}.\hskip 1em plus 0.5em minus 0.4em\relax Springer, 2014, pp.
  111--125.

\bibitem{naseer2018indoor}
M.~Naseer, S.~Khan, and F.~Porikli, ``Indoor scene understanding in 2.5/3d for
  autonomous agents: A survey,'' \emph{IEEE access}, vol.~7, pp. 1859--1887,
  2018.

\bibitem{li2020image}
P.~Li and W.~Zhao, ``Image fire detection algorithms based on convolutional
  neural networks,'' \emph{Case Studies in Thermal Engineering}, vol.~19, no.
  100625, 2020.

\bibitem{gaur2020video}
A.~Gaur, A.~Singh, A.~Kumar, A.~Kumar, and K.~Kapoor, ``Video flame and smoke
  based fire detection algorithms: A literature review,'' \emph{Fire
  Technology}, vol.~56, no.~5, pp. 1943--1980, 2020.

\bibitem{simonyan2014very}
K.~Simonyan and A.~Zisserman, ``Very deep convolutional networks for
  large-scale image recognition,'' \emph{ICLR}, 2015.

\bibitem{he2016deep}
K.~He, X.~Zhang, S.~Ren, and J.~Sun, ``Deep residual learning for image
  recognition,'' in \emph{CVPR}, 2016, pp. 770--778.

\bibitem{tan2019efficientnet}
M.~Tan and Q.~Le, ``Efficientnet: Rethinking model scaling for convolutional
  neural networks,'' in \emph{International Conference on Machine
  Learning}.\hskip 1em plus 0.5em minus 0.4em\relax PMLR, 2019, pp. 6105--6114.

\bibitem{ahmed2012bayesian}
N.~R. Ahmed, E.~M. Sample, and M.~Campbell, ``Bayesian multicategorical soft
  data fusion for human--robot collaboration,'' \emph{IEEE Transactions on
  Robotics}, vol.~29, no.~1, pp. 189--206, 2012.

\bibitem{huang2017densely}
G.~Huang, Z.~Liu, L.~Van Der~Maaten, and K.~Q. Weinberger, ``Densely connected
  convolutional networks,'' in \emph{CVPR}, 2017, pp. 4700--4708.

\bibitem{selvaraju2017grad}
R.~R. Selvaraju, M.~Cogswell, A.~Das, R.~Vedantam, D.~Parikh, and D.~Batra,
  ``Grad-cam: Visual explanations from deep networks via gradient-based
  localization,'' in \emph{ICCV}, 2017, pp. 618--626.

\bibitem{goldberg2017neural}
Y.~Goldberg, ``Neural network methods for natural language processing,''
  \emph{Synthesis lectures on human language technologies}, vol.~10, no.~1,
  p.~92, 2017.

\bibitem{pennington2014glove}
J.~Pennington, R.~Socher, and C.~D. Manning, ``Glove: Global vectors for word
  representation,'' in \emph{Proceedings of the conference on empirical methods
  in natural language processing}, 2014, pp. 1532--1543.

\bibitem{mikolov2013efficient}
T.~Mikolov, K.~Chen, G.~Corrado, and J.~Dean, ``Efficient estimation of word
  representations in vector space,'' \emph{ICLR}, 2013.

\bibitem{eder2014towards}
K.~Eder, C.~Harper, and U.~Leonards, ``Towards the safety of human-in-the-loop
  robotics: Challenges and opportunities for safety assurance of robotic
  co-workers','' \emph{International Symposium on Robot and Human Interactive
  Communication}, pp. 660--665, 2014.

\end{thebibliography}

\end{document}